\documentclass{article}
\usepackage{ijcai26}

\usepackage{times}
\usepackage{soul}
\usepackage{url}
\usepackage[hidelinks]{hyperref}
\usepackage[utf8]{inputenc}
\usepackage[small]{caption}
\usepackage{graphicx}
\usepackage{amsmath}
\usepackage{amsthm}
\usepackage{booktabs}
\usepackage{algorithm}
\usepackage{algorithmic}
\usepackage[switch]{lineno}

\usepackage{caption}
\usepackage{newfloat}
\usepackage{listings}
\usepackage{xcolor}

\title{One4Many-StablePacker: An Efficient Deep Reinforcement Learning Framework for the 3D Bin Packing Problem}

\author{
Lei Gao$^1$,
Shihong Huang$^{2,4}$,
Shengjie Wang$^{2,4}$,
Hong Ma$^{2,4,}$\thanks{
Corresponding authors. 
},
Feng Zhang$^1$,
Hengda Bao$^1$,
Qichang Chen$^1$,
Weihua Zhou$^{3,4,*}$\\
\affiliations
$^1$S.F. Technology Co., Ltd., 518054, Shenzhen, China\\
$^2$Polytechnic Institute, Zhejiang University, 310015, Hangzhou, China\\
$^3$School of Management, Zhejiang University, 310058, Hangzhou, China\\
$^4$Zhejiang Key Laboratory of Decision Intelligence, Hangzhou, China\\
\emails
hongma@zju.edu.cn, 
larryzhou@zju.edu.cn
}

\begin{document}

\maketitle 

\begin{abstract}
The three-dimensional bin packing problem (3D-BPP) is widely applied in logistics and warehousing. However, existing learning-based approaches often neglect stability constraints and struggle to generalize across diverse bin dimensions. To address this, we propose a novel deep reinforcement learning framework, One4Many-StablePacker (O4M-SP). The primary advantage of O4M-SP lies in its ability to handle variable bin dimensions in a single training process while explicitly enforcing two types of practical stability constraints: support constraints, which ensure an item's bottom center lies within the convex hull of the underlying contact area, and weight constraints, which restrict the total vertical load on an item to its bearing capacity. Our training method introduces two innovative mechanisms. First, a weighted reward function integrates the loading rate with a novel height difference metric for packing layouts, promoting improved bin utilization via flatter packing configurations. Second, clipped policy gradient optimization with tailored policy drifting mitigates entropy collapse, encouraging exploration at critical decision nodes during packing to prevent premature convergence. Extensive experiments demonstrate that O4M-SP generalizes effectively across diverse bin dimensions and significantly outperforms baseline methods. Furthermore, the framework exhibits strong practical applicability by solving complex scenarios under strict stability constraints.
\end{abstract}

\section{Introduction}
The three-dimensional bin packing problem (3D-BPP) is a classic combinatorial optimization problem (COP) aiming to pack a set of items with known dimensions into a bin to maximize space utilization while satisfying geometric feasibility constraints, such as non-overlapping and fully enclosed placements \cite{np_hard_fowler1981optimal}. Existing approaches, including exact and heuristic algorithms \cite{exact_alg_chen1995,exact_alg_martello2000three,ga_wu2010three,tabu_crainic2009}, face scalability and quality limitations when applied to large instances \cite{wu2023machine_Literature_review}. Recent efforts have applied deep reinforcement learning (DRL) to develop packing solutions from data or interactions, demonstrating superior empirical performance compared to heuristic rules \cite{table_2021attend2pack}. However, most DRL-based studies focus solely on maximizing volume utilization, neglecting stability-related constraints that are critical for real-world logistics, such as weight stability (preventing heavy-over-light stacking) and base support (ensuring sufficient contact area) \cite{zhou2024efficient}. These constraints, essential for generating feasible solutions, remain underexplored in prior work \cite{wu2023machine_Literature_review}. Moreover, most existing DRL models are designed for fixed bin sizes (see Table \ref{tab:DRL_BPP}), requiring retraining when bin dimensions vary, which limits their practical deployment in logistics with diverse bin specifications \cite{xiong2024gopt}.

To address these limitations, we propose the One4Many-StablePacker (O4M-SP) framework. The term ``One4Many" denotes the model’s capability to generalize across bins with varying dimensions---including those unseen during training---without requiring retraining. Meanwhile, ``Stable" highlights the explicit integration of stability constraints to ensure real-world feasibility. Simultaneously enforcing stability constraints while requiring cross-dimension generalization significantly increases problem complexity. Furthermore, existing DRL methods frequently struggle with single-objective reward functions and are prone to premature policy entropy collapse \cite{entropy_8020}. To mitigate these issues, we develop a new and efficient DRL-based framework. Specifically, we construct a weighted reward function that combines the loading rate with a novel height difference metric, and we implement clipped policy gradient optimization with policy drifting to promote exploration at critical decision nodes.

The main contributions are summarized as follows. First, we propose O4M-SP, the first DRL framework for offline 3D-BPP that simultaneously addresses stability constraints and generalizes across diverse bin dimensions in a single training process. Second, we design a new weighted reward function to mitigate the limitations of single-reward functions, and conduct policy entropy control at critical decision nodes to promote exploration and enhance policy quality. Finally, we demonstrate that O4M-SP achieves superior packing performance over baseline methods and confirm that O4M-SP effectively enforces practical stability constraints.

\section{Related Work} \label{Literature}

\subsection{DRL for solving COPs}

Combinatorial optimization is a key research topic in operations research and computer science \cite{schrijver2003combinatorial}, playing essential roles in path planning \cite{veres2019deep,li2024mathematical,li2025collaborative}, resource allocation \cite{huang2025balancing}, and task scheduling \cite{dolgui2019scheduling}. Current methods for solving COPs include traditional algorithms, deep reinforcement learning (DRL), and hybrid approaches. Due to their NP-hard nature, exact methods ensure optimality but are computationally inefficient, while heuristics improve speed yet face limitations in large-scale, real-time scenarios. DRL has recently demonstrated advantages in solving COPs by modeling discrete optimization tasks as sequential decision-making processes \cite{wu2024neural}, enabling efficient strategies to be learned through interactions between policy networks and the environment. Specifically, neural networks generate strategies trained via supervised learning \cite{fu2021generalize} or reinforcement learning \cite{bello2016neural}. Early work focused on classic graph optimization problems. Khalil et al.~\shortcite{khalil2017learning} integrated graph embeddings with Q-learning to achieve near-optimal solutions and strong generalization for minimum vertex cover and maximum cut problems. Later, Luo et al.~\shortcite{luo2023neural} introduced a Light Encoder and Heavy Decoder (LEHD) Transformer that dynamically captures node relationships, achieving effective generalization from small TSP and CVRP instances to solve large-scale problem instances with up to $1000$ nodes. Regarding hybrid methods, Feng and Yang~\shortcite{feng2025sorrel} proposed the Suboptimal-demonstration-guided Reinforcement Learning framework, which outperforms prior learning-based methods for mixed integer linear programming problems in both branching quality and training efficiency.

\subsection{DRL for solving 3D-BPP}
Recent advances in DRL have demonstrated strong potential for solving 3D-BPP. In Table \ref{tab:DRL_BPP}, we summarize recent learning-based approaches for solving offline 3D-BPP. 

\begin{table*}[t]
\centering
\setlength{\tabcolsep}{5.5pt}
\small
\begin{tabular}{@{}lcccccc@{}}
\toprule
\textbf{Approach} & \textbf{Bin Size} & \textbf{Module} & \textbf{State} & \textbf{Reward} & \textbf{Stability} & \textbf{RL} \\
\midrule
Hu et al.~\shortcite{table_2017} & -- & PN & item size & terminal (surf.) & none & PG+BS \\
Duan et al.~\shortcite{table_2018} & -- & PN & item size & terminal (surf.) & none & PPO \\
Laterre et al.~\shortcite{table_2019} & -- & DNN & item & terminal (surf.) & center supp. & MCTS \\
Goyal and Deng~\shortcite{table_2020_1} & Single & CNN & voxel & step. (LR) & none & PPO \\
Hu et al.~\shortcite{table_2020_2} & Single & CNN+Attn. & HM+size & terminal (util.+stab.) & conv. hull & AC \\
Li et al.~\shortcite{table_2020_3} & Multi. & Attn. & item & step. (LR) & none & AC \\
Jiang et al.~\shortcite{table_2021_1} & Single & CNN & HM+item & step. (LR) & none & A2C \\
Zhang et al.~\shortcite{table_2021attend2pack} & Multi. & CNN+Attn. & item+frontier & terminal (util.) & none & PO \\
Jiang et al.~\shortcite{table_2021solving} & Single & CNN+Attn. & HM+item & step. (LR) & none & A2C \\
Zhao et al.~\shortcite{table_2024dynamic} & Single & CNN+Attn. & HM+item & step. (LR) & none & SAC \\
Wang et al.~\shortcite{table_2025bin} & Single & PN+GRU+Attn. & HM+item & step. (LR+comp.+pyr.) & none & AC \\
\midrule
\textbf{Our Work} & \textbf{Multi.} & \textbf{Attention} & \textbf{Bin+item} & \textbf{step. (LR+HD)} & \textbf{supp.+weight} & \textbf{PPO+Entropy} \\
\bottomrule
\end{tabular}
\caption{Comparison of DRL-based approaches for the offline 3D-BPP. Abbreviation: Point Network (PN), Policy Gradient (PG), Beam Search (BS), Proximal Policy Optimization (PPO), Deep Neural Network (DNN), Monte Carlo Tree Search (MCTS), Convolutional Neural Networks (CNN), Actor-Critic (AC), Advantage Actor-Critic (A2C), Prioritized Oversampling (PO), Soft Actor-Critic (SAC), Gated Recurrent Unit (GRU), Attention (Attn.), Height Map (HM), Surface (Surf.), Stepwise (Step.), Utilization (Util.), Loading Rate (LR), Support (Supp.), Compactness (Comp.), Pyramid (Pyr.), Height Difference (HD)}
\label{tab:DRL_BPP}
\end{table*}

To our knowledge, Hu et al.~\shortcite{table_2017} first introduced a pointer network with policy gradient to minimize bin surface area for a novel variant of 3D-BPP, laying the groundwork for sequential decision frameworks for 3D-BPP. Subsequent efforts focused on model and algorithmic improvements. The approach in Duan et al.~\shortcite{table_2018} applied the more stable PPO algorithm, whereas Laterre et al.~\shortcite{table_2019} integrated deep neural networks (DNNs) with Monte Carlo Tree Search (MCTS) and introduced center-of-mass support constraints. In a related line of work, search-based methods such as the Packing Configuration Tree (PCT) have been proposed to explore continuous action spaces for bin packing tasks~\cite{zhao4}. With the rise of convolutional neural networks (CNNs), height map-based state representations became prevalent \cite{table_2020_1,table_2021_1,table_2021solving,table_2024dynamic}. Height maps provide a compact encoding of space occupancy and, when paired with CNNs, deliver strong performance for fixed-size bins. However, the fixed input structure of CNNs restricts their generalization to variable-sized bins, limiting practical applicability. Recent studies have aimed to address this limitation: Zhang et al.~\shortcite{table_2021attend2pack} leverage self-attention to model inter-item and item-bin relationships directly, demonstrating promise in handling variable-sized bins. The approach in Hu et al.~\shortcite{table_2020_2} combines CNNs (height maps) with attention mechanisms (for item features), introducing a stability ratio objective function and ``convex hull checks", marking an early effort to integrate spatial efficiency with stability. For a comprehensive overview of the evolution of DRL in bin packing problems, readers are referred to Dahmani et al.~\shortcite{DAHMANI2025100616}.

Despite the importance of stability in real-world logistics, our survey reveals that existing DRL-based offline 3D-BPP research has largely neglected this constraint. As detailed in Table~\ref{tab:DRL_BPP}, exceptions are limited: Laterre et al.~\shortcite{table_2019} introduced center-of-mass support constraints, while Hu et al.~\shortcite{table_2020_2} implemented a simplified stability check through convex hull checks. In contrast, physical stability remains a central requirement in the adjacent field of robotic packing \cite{zhao1,zhao2}. Notably, progress has been made in online 3D-BPP, where bottom support constraints have been considered and modeled \cite{yang2023heuristics,zhao2021online,zhou2024efficient}. These studies offer valuable precedents for integrating complex yet practical stability constraints into offline 3D-BPP.  

\section{Methodology}

\subsection{Problem Statement and Formulation}
In this subsection, we formulate 3D-BPP with stability constraints. 
The problem studied aims to find a non-overlapping placement of $N$ cuboid items into a bin. Each item $i$ has dimensions $(l_i,w_i,h_i)$, where $l_i$, $w_i$, and $h_i$ represent its length, width, and height, respectively. Similarly, the bin has fixed dimensions ($L$, $W$, $H$), where the maximum possible height is $H_N=\sum_{i=1}^{N}\max(l_i,w_i,h_i)$. The objective is to maximize space utilization $\eta$:

\begin{equation}
    \eta = \frac{\sum_{i=1}^{N} w_i l_i h_i}{W L H_N},
\end{equation}

$H_N$ is the minimal bin height required to enclose all items. 
The $X$, $Y$, and $Z$ axes correspond to the bin’s length, width, and height, respectively. All items must be axis-aligned, with edges parallel to the axes, and may be placed with any face as the base. Beyond these geometric constraints, we  incorporate stability constraints (Figure \ref{Total_Methods}(b)), comprising two main parts:

\textbf{Support Constraints}:
To ensure physical stability, a placed item must be adequately supported by the items beneath it. Specifically, we compute the convex hull of the contact areas formed between the new item's base and the supporting items. The constraint is satisfied if and only if the geometric center of the item's bottom surface lies within this convex hull; otherwise, the placement is considered unstable.

\textbf{Weight Constraints}: 
These enforces that the vertical load on any item does not exceed its bearing capacity. For a new item $j$, the weight contribution $G_{ji}$ transmitted to a supporting item $i$ is calculated based on the contact area ratio. The placement is feasible only if $G_{cur}^{(i)}+G_{ji} \leq r_w G_i$ for all supporting items, where $G_{cur}^{(i)}$ is the existing load on item $i$ and $r_w$ is the maximum load-bearing ratio relative to the item's own weight $G_i$.

We formulate the above 3D-BPP as a Markov Decision Process (MDP), which is a tuple $<S, A, P, R, \gamma>$. 

\textbf{State}: At time step $t$, the environment state is defined as $s_t = \{s_{bin}, s_{valid\_items}\}$. First, for the bin state $s_{bin}$, existing methods have limitations: height maps \cite{table_2024dynamic} and weighted 3D voxel grids \cite{yang2023heuristics}, which rely on CNNs, limit the model’s ability to generalize across diverse bin dimensions; the item list representation \cite{zhao2021learning} not fully capture remaining space; and the Empty Maximal Space (EMS)-based PG representation \cite{xiong2024gopt} captures all available space but provides a weak representation of packed items. To address these issues, we propose a hybrid state representation that combines the strengths of these approaches. At time step $t$, with $n$ items packed, $s_{bin}$ is represented as an $(n+2) \times 7$ matrix: 
The first row contains the bin’s attributes. The second row contains the attributes of the current space. The subsequent $n$ rows represent the attributes of the packed items. A right-handed coordinate system is used with the front-left-bottom (FLB) vertex of the bin as the origin; columns 1–3 of the matrix represent the FLB coordinates, with the bin’s coordinates being $(0,0,0)$, columns 4–6 represent the length, width, and height, and column 7 represents the weight, with the weights of the bin and the current space being 0. The current space (highlighted in orange in Figure~\ref{Total_Methods}(a)) is selected from all available spaces (highlighted in blue $S_1$ to $S_5$ in Figure~\ref{Total_Methods}(a)) generated by the Empty Maximal Space (EMS) method using a screening heuristic. The screening heuristic is as follows: The loading sequence follows an XYZ order (prioritizing the X-axis, then the Y-axis, and finally the Z-axis) to maximize floor and space utilization. In our implementation, all available spaces are pushed onto a stack.
We evaluate the stability of the space at the top of the stack; if valid, it is selected as the current target space. If unstable, the top space is popped from the stack, and the next top is evaluated.

Second, the representation of the valid items state $s_{valid\_items}$ is closely tied to the stability check. At time step $t$, if no available space exists, no valid item or valid action is available. Otherwise, the stack containing available spaces is checked as follows: the top space is evaluated, and the remaining items and their rotations are examined to identify valid orientations that can be placed in the space. If no valid orientations exist, the top space is popped from the stack, and the next top space is evaluated; if valid orientations are found, a stability check is performed on the candidate actions. Actions that pass the stability check are included in $s_{valid\_items}$, and the top space in the stack is selected as the current space. At time step $t$, with $k$ types of valid items, the state $s_{valid\_items}$ is represented as a $k \times 5$ matrix: columns 1–3 represent the length, width, and height of each item, column 4 represents the remaining quantity of each item, and column 5 represents the weight of each item.

\textbf{Action}: Given the state $s_t$, action $a_t$ selects a valid item and its rotation, placeing it at the front-left-bottom (FLB) corner of the current target space. The dimension of the action space depends on the number of valid items and their feasible rotations.

\textbf{State Transition}: State transitions are deterministic. Specifically, given the state $s_t$, action $a_t$ is sampled according to the policy $\pi(\cdot|s_t)$, transitioning deterministically to the state $s_{t+1}$.

\textbf{Reward}: An effective stepwise reward provides meaningful intermediate signals to guide policy exploration. Existing methods employ mainly a single reward, such as volume change \cite{li2022one}, volume gap reduction \cite{table_2020_3}, loading-rate change \cite{zhang2024online}, or gap-ratio reduction \cite{table_2024dynamic}, which may lead policies to converge prematurely to local optima. In this work, we introduce a new weighted reward (WR):
\begin{equation}
\label{reward_function_eqa}
{r_t} = \alpha_1 r_t^{_{LR}} + \alpha_2 r_t^{HD}
\end{equation}

where $r_t^{LR}$ denotes the loading rate reward (Eq. \ref{loading_rate_eqa}), $r_t^{HD}$ denotes the height difference reward for the packing layout (Eq. \ref{height_difference_eqa}), and $\alpha_1$, $\alpha_1$ are hyperparameters.
\begin{equation}
\label{loading_rate_eqa}
r_t^{_{LR}} = \frac{{\sum\nolimits_{i = 1}^t {{w_i}{l_i}{h_i}} }}{{WL{H_t}}} - \frac{{\sum\nolimits_{i = 1}^{t - 1} {{w_i}{l_i}{h_i}} }}{{WL{H_{t - 1}}}}
\end{equation}

\begin{equation}
\label{height_difference_eqa}
r_t^{HD} = ({H_t} - {H_t}^\prime ) - ({H_{t - 1}} - {H_{t - 1}}^\prime )
\end{equation}

\begin{figure}[t]
\centering
\includegraphics[width=\columnwidth]{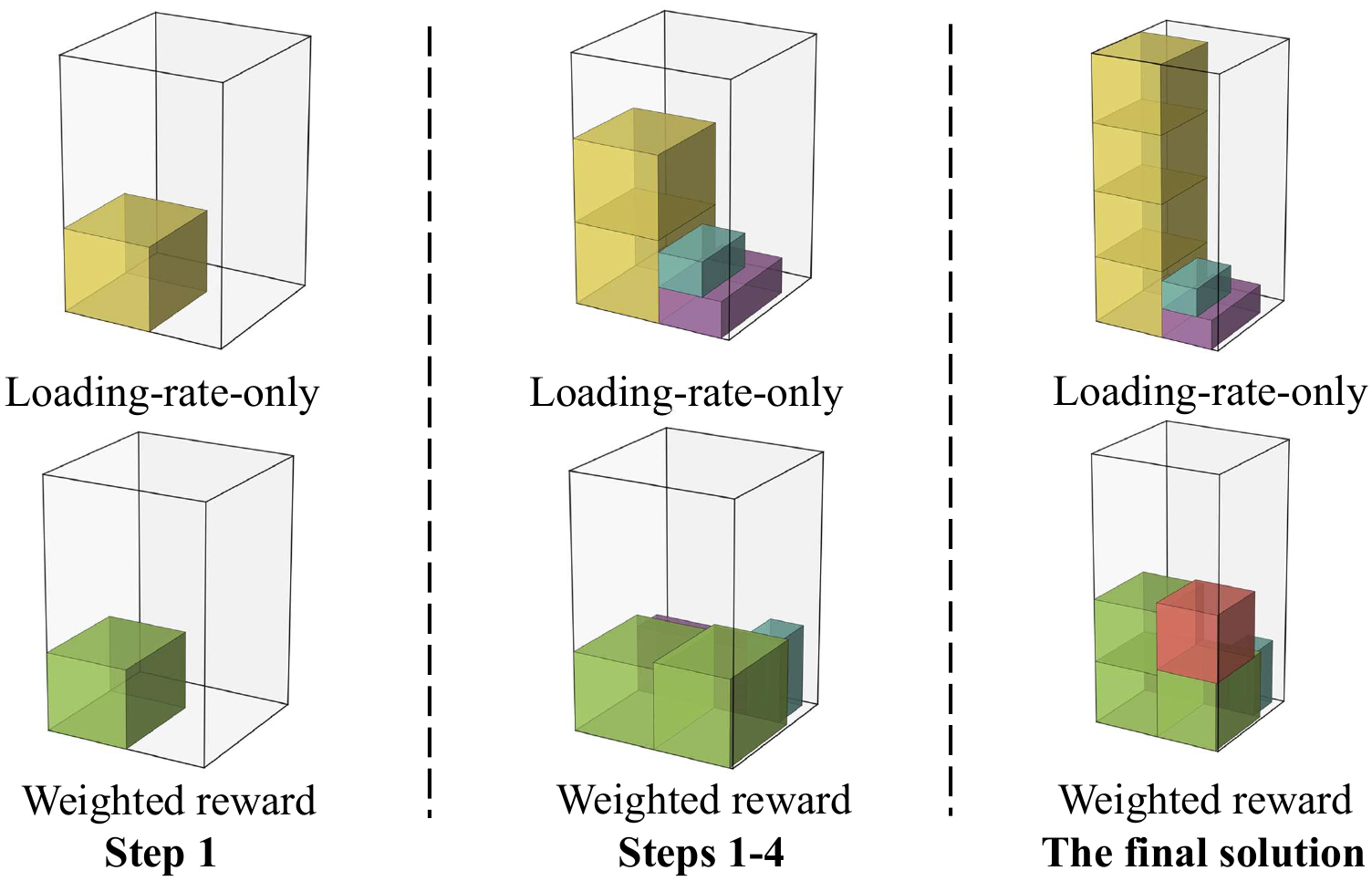} %
\caption{
The loading-rate-only reward induces greedy vertical stacking, leading to surface fragmentation and restricted placement options (top). In contrast, the weighted reward (WR) promotes surface flatness, preserving a larger feasible action space and preventing premature convergence to local optima (bottom).}
\label{height_difference}
\end{figure}

In Equations~(\ref{loading_rate_eqa})-(\ref{height_difference_eqa}), $H_t$ denotes the maximum height of all packed items at step $t$, and ${H_t}^\prime$ denote the second maximum height of all packed items. We normalize the difference between $H_t$ and ${H_t}^\prime$, assigning rewards to smaller height difference metrics to encourage flatter packing (See Figure \ref{height_difference}).

We theoretically validate the proposed reward mechanism. Detailed proofs are provided in our GitHub project~\footnote{\url{https://github.com/chaojihetao/3dbpp}}. First, we establish that the stepwise loading rate reward $r_t^{LR}$ forms a telescoping sum, meaning the cumulative reward exactly equals the final global utilization $\eta$. 
This property ensures that maximizing the stepwise reward is equivalent to maximizing the global packing objective. Next, regarding the height difference reward $r_t^{HD}$, we employ Potential-Based Reward Shaping (PBRS) \cite{ng1999policy} with the potential function $\Phi(s)$ representing surface flatness: 
\begin{equation}
\Phi(s) = - \beta \cdot (H^{(1)}(s) - H^{(2)}(s)), 
\end{equation}
where $H^{(1)}$ and $H^{(2)}$ denote the maximum and second-maximum heights, respectively. 
Since $\Phi(s)$ depends solely on the state, this shaping term guarantees policy invariance, accelerating convergence by encouraging flat packing surfaces without altering the optimal policy $\pi^*$. Finally, we demonstrate that minimizing height variance maximizes the measure of the feasible action space. 
By explicitly penalizing height variance, the agent is guided to reduce surface fragmentation, thereby preserving larger continuous support areas for subsequent placements. 

\subsection{Network Architecture}
The network architecture significantly influences the learning and generalization capabilities of DRL algorithms. To address limitations of existing approaches, we propose a novel state representation that enables model generalization across bins of varied dimensions and introduce a stability checker module to ensure adherence to physical stability constraints. The model adopts an actor-critic framework, as shown in Figure~\ref{Total_Methods}.
The packing environment provides the bin state, valid item candidates, and rotation information, which are encoded into 256-dimensional embeddings. These embeddings are processed by the feature extractor, actor, and critic networks to produce action probabilities and value estimates, respectively. The network architecture comprises the following core components:

\begin{figure*}[t]
\centering
\includegraphics[width=2.0\columnwidth]{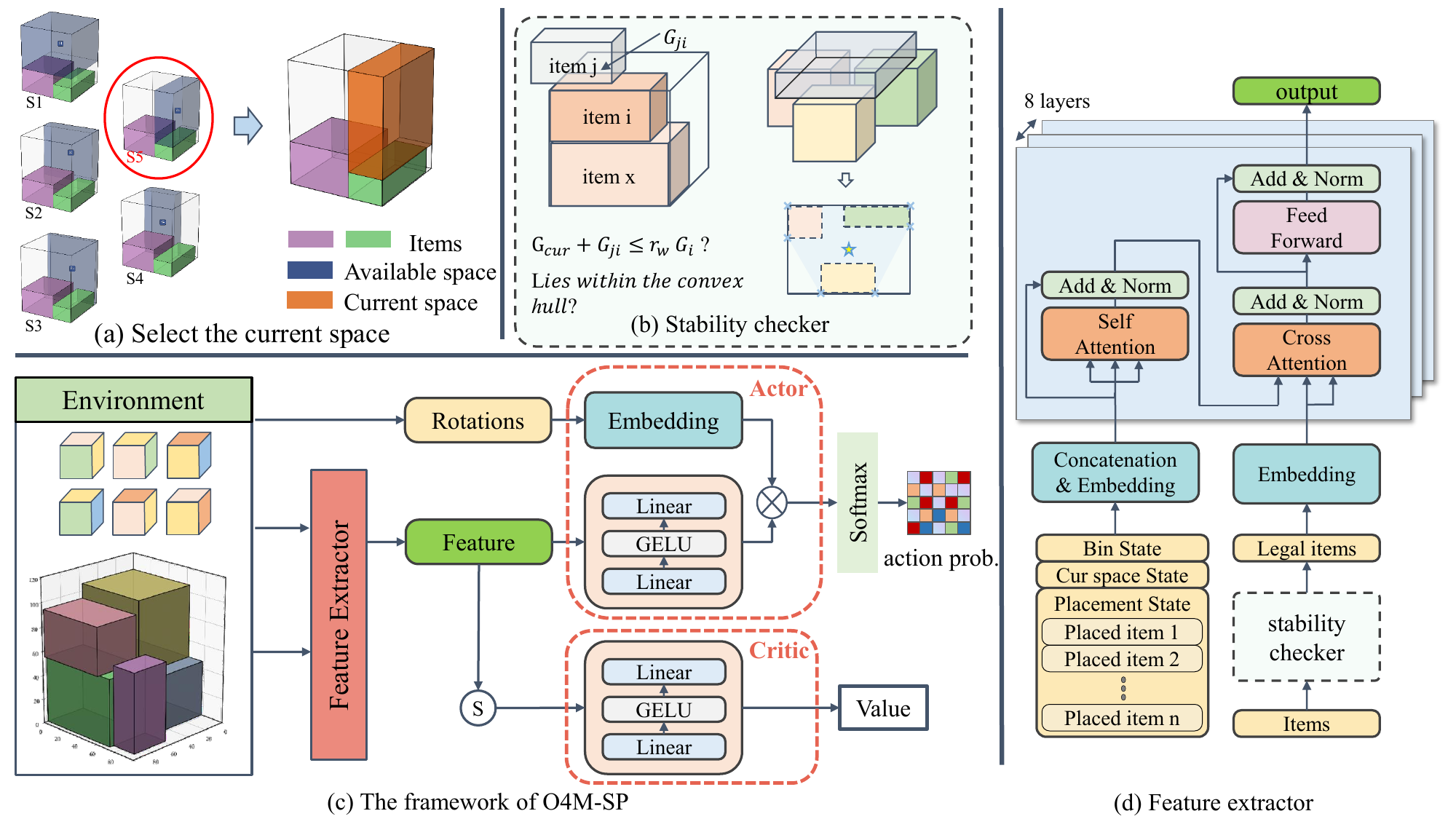}
\caption{Overview of our method. (a) Space selection: A heuristic for selecting the space for placement. (b) Stability checker: Enforces support and weight constraints. (c) O4M-SP framework: Extracts features from environmental information and uses actor and critic networks to generate action probabilities and value estimates. (d) Feature extractor: Eight identical encoder blocks, each comprising self-attention, cross-attention, feed-forward layers, residual connections, and layer normalization.}
\label{Total_Methods}
\end{figure*}

\textbf{Feature Extractor}: The feature extractor consists of eight identical encoder blocks, each comprising a self-attention layer, a cross-attention layer, and a feed-forward network (FFN), followed by residual connections and layer normalization. The self-attention layer integrates information from the bin, current space, and placed items to produce an updated state vector $S_{new}$. Along with $S_{valid\_items}$, it is fed into and processed by the cross-attention layer to enable inter-feature interactions. The FFN in each block performs nonlinear transformations on the attention outputs, enhancing feature expressiveness.

\textbf{Actor and Critic Networks}: Both are two-layer multilayer perceptrons  (MLPs) with GELU activation functions. The actor network processes state and rotation features separately, then combines them multiplicatively to produce a probability distribution over actions. The critic network directly estimates the state value from the packing state features.

\subsection{Training Method}
A core challenge in training DRL neural networks lies in the exploration-exploitation trade-off \cite{drl_1988}. Policy entropy, an indicator of exploration capability reflecting action diversity, typically decreases as model performance improves \cite{entropy_cov}. 

\begin{figure}[t]
\centering
\includegraphics[width=\columnwidth]{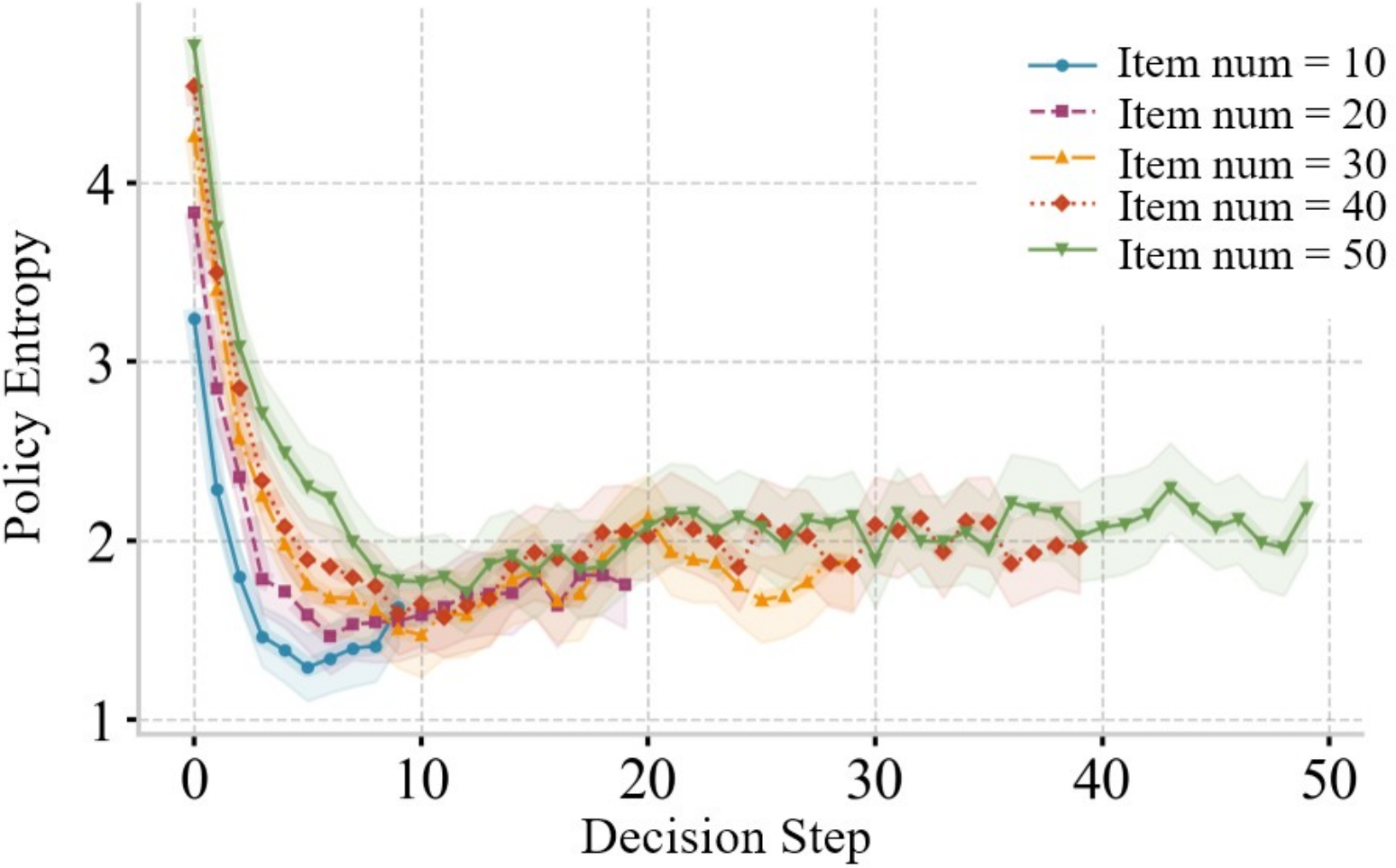} %
\caption{Variation of policy entropy with decision steps}
\label{entropy_first_step}
\end{figure}

To address this, we propose a tailored entropy-control scheme integrated into PPO for 3D-BPP. 
First, we note that prior studies identify critical decision nodes where policy entropy decreases significantly, narrowing the action space \cite{entropy_8020}. Accordingly, we selectively apply entropy control at critical decision nodes during packing to restrict high-confidence decisions and enhance the model’s ability to escape local optima. We observe that in 3D-BPP, entropy consumption is most pronounced during the initial steps, with entropy curves for varying item quantities shown in Figure~\ref{entropy_first_step}. 

Next, as indicated by Cui et al.~\shortcite{entropy_cov}, when policy parameters are updated from $\theta^{k}$ to $\theta^{k+1}$, the change in policy entropy $H$ for a given state $s$ follows the pattern presented in Equation~(\ref{cui}).
\begin{equation}
\label{cui}
  \Delta H \approx   -\beta Cov_{a \sim \pi_\theta^{k}(\cdot \mid s)} \bigl(\,\log \pi_\theta^{k}(a \mid s), A(s,a)\bigr)
\end{equation}
where $\Delta H = H(\pi_\theta^{k+1} \mid s) - H(\pi_\theta^{k} \mid s)$  and $H(\pi_\theta^{k} \mid s) = -\sum_{a} \pi_\theta^{k}(a \mid s)\, \log \pi_\theta^{k}(a \mid s)$ represents the policy entropy at step $k$ for state $s$, $A(s,a)$ denotes the advantage of action $a$, and ${Cov}_{a \sim \pi_\theta^{k}(\cdot \mid s)}(\cdot,\cdot)$ denotes the covariance calculated for all possible actions $a$ under the policy $\pi_\theta^{k}(\cdot \mid s)$ at step $k$. It reveals that when the covariance at step $k$ is high, forward exploration leads to significant policy entropy loss, implying that high-covariance nodes are critical decision nodes. 

Based on these findings, we conduct policy entropy control on the above two types of critical decision nodes. For nodes $N_{high\_cov}$ with high covariance, we extract a small portion $\phi \cdot N_{high\_cov}$ and apply the clipped policy gradient optimization proposed in Cui et al.~\shortcite{entropy_cov}:
\begin{equation}
{p_t}(\theta ) = \left\{ {\begin{array}{*{20}{c}}
{\frac{{{\pi _\theta }({a_t}|{s_t})}}{{{\pi _{{\theta _{old}}}}({a_t}|{s_t})}},  \hspace{1cm} t \notin \phi {N_{high\_cov}}}\\
{0, \hspace{2.3cm} t \in \phi {N_{high\_cov}}}
\end{array}} \right.
\end{equation}

This approach enables the model to explore high-covariance nodes while restricting certain nodes to perform only forward propagation, effectively mitigating the neural network’s overfitting and preserving policy entropy. 
However, for the nodes $N_{\mathrm{first\_step}}$ corresponding to initial item placements, clipping may adversely affect the model’s convergence. Hence, we apply a tailored policy drifting method, similar to Liu et al.~\shortcite{liu2025cpgd}.
\begin{equation}
{J_t}(\theta ) = \left\{ {\begin{array}{*{20}{c}}
{{J_t}(\theta ) \hspace{2.85cm}t \notin {N_{first\_step}}}\\
{{J_t}(\theta ) - \beta {P_{policy}}, \hspace{1cm} t \in {N_{first\_step}}}
\end{array}} \right.
\end{equation}

$J_{p}(\theta)$ denotes the policy loss term, which includes a penalty $P_{policy} = |log(\frac{\pi_{\theta}(a_{t} \mid s_{t})}{\pi_{\theta_{old}}(a_{t} \mid s_{t})})|$. The method mitigates policy deviation during initial item placement, weakens backpropagation intensity, and avoids premature convergence to local optima, thereby slowing policy entropy loss.

\begin{table*}[!h]
\centering
\small
\begin{tabular}{@{}ccccccccc@{}}
\toprule
\textbf{} & \textbf{GA} & \textbf{MTSL} & \textbf{CQL} & \textbf{MM} & \textbf{A2P} & \textbf{DMRL} & \textbf{O4M-SP(S)} & \textbf{O4M-SP(M)} \\
\midrule
10 & 72.80\% & 49.80\% & 72.60\% & - & - & 73.00\% & 73.41\% ± 0.0022 & \textbf{76.12\%} ± 0.0017 \\
16 & 68.50\% & 54.70\% & 69.40\% & - & - & 77.20\% &79.14\% ± 0.0011 & \textbf{79.76\%} ± 0.0011 \\
20 & 66.10\% & 57.30\% & 67.60\% & 71.80\% & 76.70\% & 78.90\% & \textbf{81.44\%} ± 0.0009 & 80.40\% ± 0.0007 \\
30 & 62.60\% & 59.40\% & 69.80\% & 75.50\% & 79.70\% & 81.10\% & \textbf{84.61\%} ± 0.0005 & 77.71\% ± 0.0012 \\
\bottomrule
\end{tabular}
\caption{Results: O4M-SP vs. Baseline methods, including: (1) Genetic Algorithm (GA) \protect\cite{ga_wu2010three}, (2) Multi-Task Selected Learning (MTSL) \protect\cite{table_2018}, (3) Conditional Query Learning (CQL) \protect\cite{table_2020_3}, (4) Multimodal Deep Reinforcement Learning (MM) \protect\cite{table_2021solving}, (5) Attend2Pack (A2P) \protect\cite{table_2021attend2pack} and (6) Dynamic Multi-modal Deep Reinforcement Learning (DMRL) \protect\cite{table_2024dynamic}}
\label{tab:baseline}
\end{table*}

\section{Experiments}
\subsection{Experimental setup}
We implement O4M-SP on an Ubuntu 24.04 Linux server equipped with dual AMD EPYC 9554 64-Core Processors, 768GB memory, and an NVIDIA L40S GPU with 48GB memory. Model training takes about 37 hours. 
Training consists of 10,000 iterations, using the Adam with Weight Decay (AdamW) optimizer \cite{loshchilov2017decoupled} with a batch size of 256. The learning rate schedule employs a warm-up phase (20 epochs) followed by cosine annealing, with an initial learning rate of $10^{-4}$ and a minimum learning rate of $10^{-6}$. Performance is evaluated every 20 iterations on an independent validation set of 10,000 samples, and the best model is saved. Training incorporates an early stopping mechanism with a patience value of 100 evaluations. The source code and data generation details are publicly available at: \url{https://github.com/chaojihetao/3dbpp}.

\begin{table*}[!t]
\centering
\small
\begin{tabular}{lcccccccc}
\hline
Model & B1-30 & Gap & B2-30 & Gap & B3-30 & Gap & BM-M & Gap \\ \hline
MCTS & 84.34\% & * & 82.14\% & * & 85.73\% & * & 78.99\% & * \\
Greedy & 69.62\% & 17.45\% & 68.81\% & 16.23\% & 69.89\% & 18.48\% & 67.38\% & 14.70\% \\
Grasp & 74.04\% & 12.21\% & 73.99\% & 9.92\% & 73.68\% & 14.06\% & 74.05\% & 6.25\% \\
\textbf{O4M\_SP} & \textbf{78.72\%} & \textbf{6.66\%} & \textbf{77.91\%} & \textbf{5.15\%} & \textbf{78.70\%} & \textbf{8.20\%} & \textbf{75.13\%} & \textbf{4.89\%} \\ \hline
\end{tabular}
\caption{Generalization results: O4M-SP vs. Heuristic algorithms}
\label{tab:generalization2}
\end{table*}

\subsection{Baseline}
To evaluate O4M-SP, we select representative, publicly available algorithms from the literature as baselines. Table~\ref{tab:baseline} reports the average loading rates and their variances for 10,000 test instances after 10,000 training iterations. All experiments are conducted with a fixed bin size of $100 \times 100 \times H$, evaluating each method’s performance on instances with 10, 16, 20, and 30 items. To investigate the effect of training sets with varied bin dimensions, we train O4M-SP on two datasets: one with fixed bin size ($100 \times 100 \times H$), denoted O4M-SP(S), and another with bin dimensions independently drawn from a uniform distribution over the interval $[100, 450]$, denoted O4M-SP(M). Following Zhao et al.~\shortcite{table_2024dynamic}, all DRL-based methods, including ours, decode 128 samples per instance and retain the best solution. 

Experimental results demonstrate that O4M-SP consistently outperforms all baseline methods across varied problem instances. Notably, O4M-SP(S) and O4M-SP(M) exhibit distinct strengths depending on instance complexity. On smaller instances (10 and 16 items), O4M-SP(M) significantly outperforms both baseline methods and O4M-SP(S), indicating that small-scale packing problems are less sensitive to bin dimensions. Training with random dimensions enhances the model’s generalization ability, mitigating overfitting to fixed bin sizes. Conversely, O4M-SP(S) achieves superior performance on larger instances (20 and 30 items), suggesting that sensitivity to bin dimensions increases with problem scale. Hence, specialized training on fixed bin sizes performs better on larger problems.

\begin{table}[!h]
\centering
\small
\begin{tabular}{@{}ccccc@{}}
\toprule
Model & 10 & 16 & 20 & 30  \\
\midrule
DMRL-BPP & 73.00\% & 77.20\% & \textbf{78.90\%} & 81.10\% \\
O4M-SP (train\_10) & \textbf{75.41\%} & \textbf{78.10\%} & 77.81\% & \textbf{81.98\%} \\
\bottomrule
\end{tabular}
\caption{O4M-SP (train\_10) vs. Baseline methods}
\label{tab:generalization1}
\end{table}

\subsection{Generalization}
To evaluate O4M-SP's generalization ability, we conduct two additional experiments. First, we test varied item quantities. Similar to the data generation method in  Zhao et al.~\shortcite{table_2024dynamic}, where the bin size is fixed at $100 \times 100 \times H$ with 10 items per instance, we construct the training set and test it on benchmark sets with varied sizes (10, 16, 20, and 30 items). As shown in Table~\ref{tab:generalization1}, the model trained on the training set with 10 items generalizes well. Compared to the models in Zhao et al.~\shortcite{table_2024dynamic} trained on test sets with 10, 16, 20, and 30 items, O4M-SP improves
loading rates by 2.41\%, 0.90\%, and 0.88\% on the 10-, 16-, and 30-item test sets, respectively. O4M-SP performs slightly worse (1.09\%) on the 20-item test set.

Second, we evaluate the generalization capability of O4M-SP on bin dimensions not encountered during training. The experimental setup comprises three fixed-size test sets (B1\_30, B2\_30, B3\_30) with bin dimensions $(150, 230, 180)$, $(176, 153, 203)$, and $(298, 103, 159)$, alongside a mixed test set (BM\_M) featuring randomized bin geometries and item quantities.  Since existing learning-based methods typically fail to generalize to unseen bin dimensions without retraining, we compare O4M-SP against two heuristic algorithms: Greedy and GRASP.
Additionally, we employ MCTS (configured with an intensive search time of one hour per instance) as a high-performance baseline to quantify the performance gap. As summarized in Table~\ref{tab:generalization2}, O4M-SP demonstrates robust zero-shot generalization capability. It consistently outperforms both the Greedy and Grasp algorithms by a significant margin across all test scenarios. While the loading rate of O4M-SP is 5.18\% lower than that of MCTS, it offers superior computational efficiency: O4M-SP solves all instances in minutes, whereas MCTS requires nearly an hour per instance.

\subsection{Ablation study}
\begin{figure}[!h]
\centering
\includegraphics[width=\columnwidth]{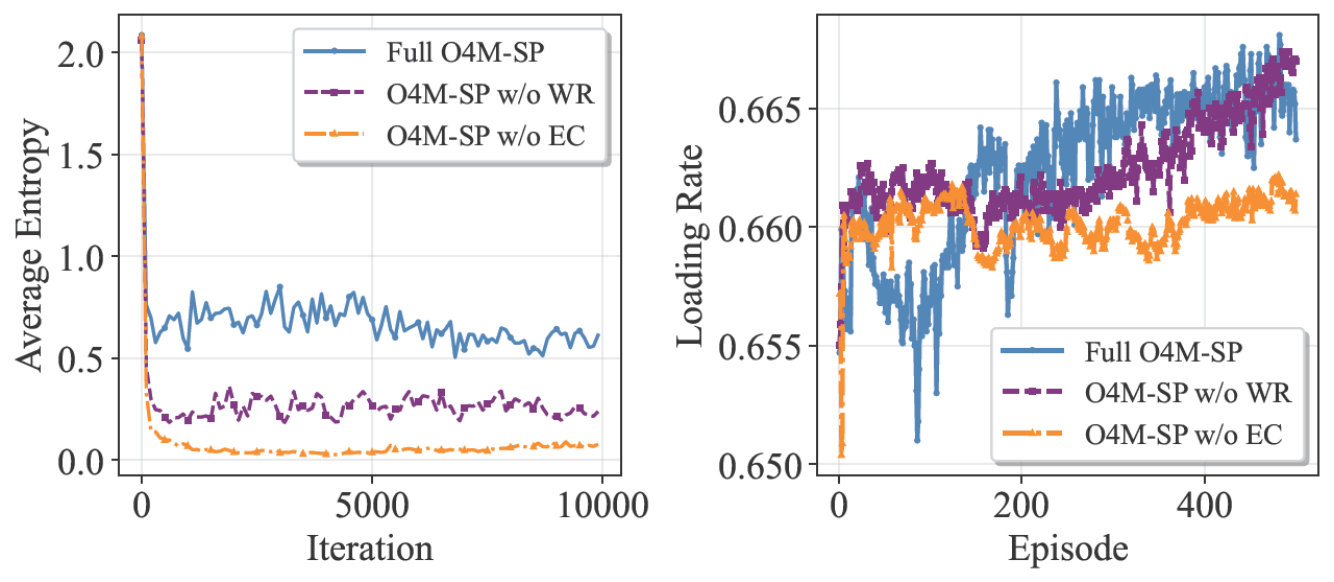} %
\caption{Policy entropy curves}
\label{ablation_figure}
\end{figure}

To assess the contributions of key components in O4M-SP, namely the weighted reward (WR) and entropy control (EC), we conduct ablation studies using three variants: O4M-SP, O4M-SP without EC,
and O4M-SP without WR.
A multi-dimensional test dataset is constructed, comprising bins with fixed dimensions (S1: $100 \times 100  \times H$; S2: $300 \times 150 \times H$) and bins with dimensions randomly drawn from a uniform distribution over [100, 450]. Bin height $H$ is set to the sum of the maximum dimensions (length, width, or height) of each items. Item quantities are either fixed (suffix \_10, \_30) or randomly generated (suffix \_M). This yields seven test sets: S1\_10, S1\_30, S2\_10, S2\_30, M\_10, M\_30, and M\_M. The ablation study results are summarized in Table~\ref{tab:ablation}. The policy entropy curves for each model during training are presented in Figure~\ref{ablation_figure}.

As illustrated in Table~\ref{tab:ablation}, the O4M-SP model consistently achieves superior performance across all test scenarios. Specifically, the average loading rate of the complete model surpasses that of O4M-SP w/o EC and O4M-SP w/o WR by 4.87\% and 5.72\%, respectively, validating the distinct contribution of each component. Figure~\ref{ablation_figure} further corroborates these findings. The average entropy curves (left) demonstrate that the EC module effectively sustains higher policy entropy a higher level of policy entropy, preventing premature convergence and maintaining exploration capability. Consequently, as shown in the loading rate curves (right), the full O4M-SP model demonstrates accelerated convergence and attains a superior final loading rate compared to its counterparts.

\begin{table}[!h]
\centering
\setlength{\tabcolsep}{0.7mm}
\small
\begin{tabular}{@{}cccccccc@{}}
\toprule
Model & \textbf{S1\_10} & \textbf{S1\_30} & \textbf{S2\_10} & \textbf{S2\_30} & \textbf{M\_10} & \textbf{M\_30} & \textbf{M\_M} \\
\midrule
O4M-SP & \textbf{66.87} & \textbf{76.73} & \textbf{67.06} & \textbf{79.31} & \textbf{66.64} & \textbf{78.49} & \textbf{74.89} \\
O4M-SP w/o EC & 62.22 & 73.22 & 63.08 & 76.76 & 61.24 & 74.88 & 70.63 \\
O4M-SP w/o WR & 62.45 & 72.59 & 61.87 & 75.65 & 61.94 & 74.97 & 72.89 \\
\bottomrule
\end{tabular}
\caption{Results (\%) of ablation study on multiple data sets}
\label{tab:ablation}
\end{table}

\subsection{Handling Stability Constraints}
\begin{figure}[!h]
\centering
\includegraphics[width=0.95\columnwidth]{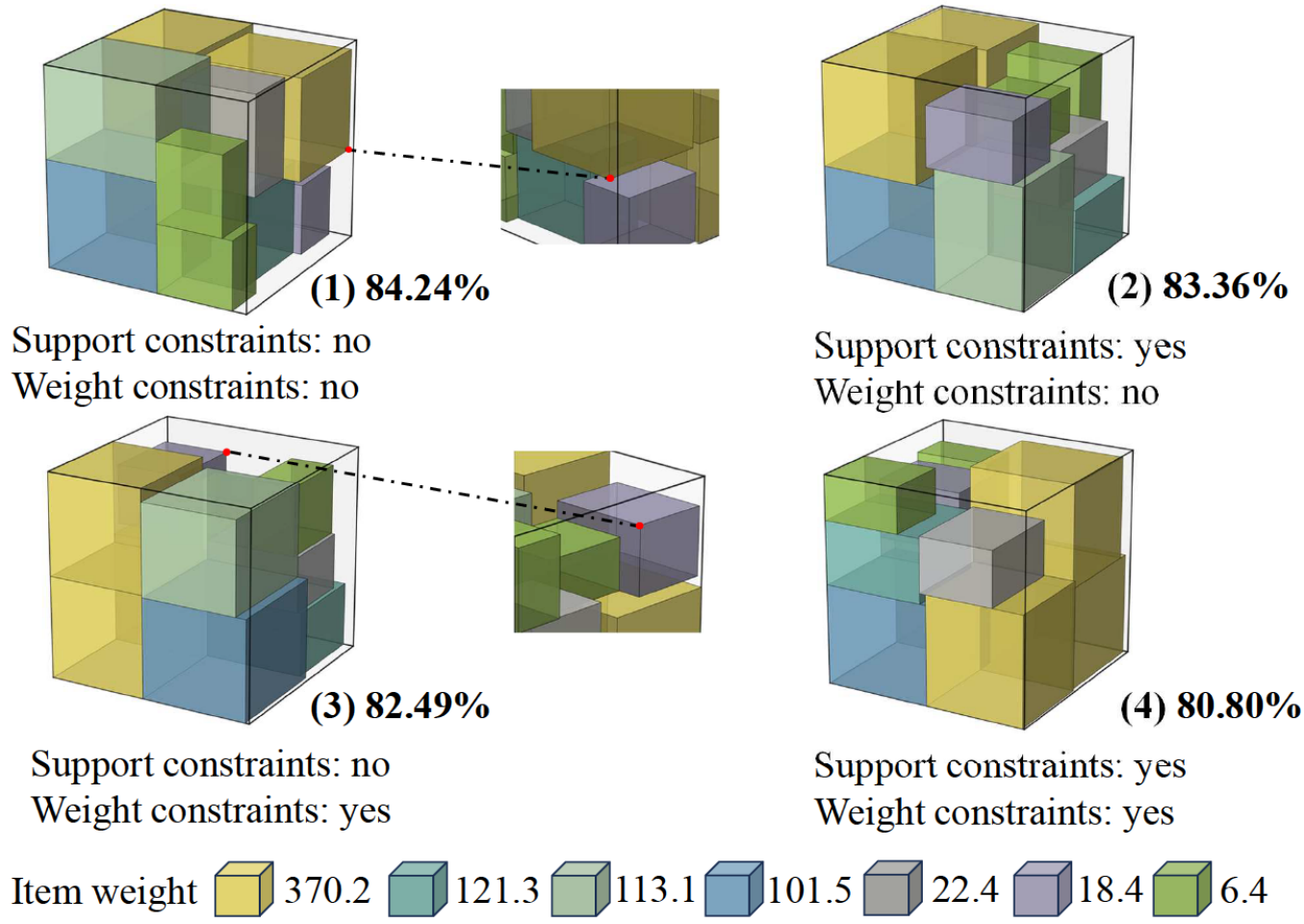} %
\caption{Visualizations for handling stability constraints}
\label{support_weight}
\end{figure}

To validate O4M-SP's ability to handle practical constraints, we conduct a case study using seven item types (see Figure~\ref{support_weight}). The bin has fixed length and width dimensions of 100, while the height is set to the maximum item height multiplied by the total number of items. 
We test four scenarios based on the definitions provided in Section 3: the support constraints mandate that an item's bottom center lies within the supporting convex hull, while the weight constraints ensure the load on any item does not exceed its load-bearing capacity. As illustrated in Figure~\ref{support_weight}, Scenario 1 (unconstrained) achieves the highest loading rate of 84.24\%, followed by Scenario 2 (support only) at 83.36\%, Scenario 3 (weight only) at 82.49\%, and Scenario 4 (both constraints) at 80.80\%. Visual inspection reveals that Scenarios 1 and 2 violate weight limits, as heavy items (yellow) are stacked atop lighter ones. Conversely, Scenarios 3 and 4 satisfy weight constraints by placing heavy items at the bottom or stacking them on identical items. Similarly, Scenarios 2 and 4 successfully enforce support constraints, ensuring sufficient base contact for all items. In contrast, Scenarios 1 and 3 exhibit support failures, characterized by item centers falling outside the underlying contact convex hull. These results confirm O4M-SP’s effectiveness in strictly enforcing stability constraints, underscoring its suitability for practical industrial applications.

\section{Conclusion}
We present a novel deep reinforcement learning framework, One4Many-StablePacker (O4M-SP), which integrates complex yet practical stability constraints to enhance real-world applicability. A core advantage is its ``train once, apply broadly" capability, which enables generalization across diverse bin dimensions 
Extensive experiments demonstrate O4M-SP's superior performance over existing methods across various problem instances. Future research could extend the framework to address irregularly shaped items and refined load distribution models for multi-support configurations, thus enabling more comprehensive applications.

\clearpage
\section*{Acknowledgments}
This work is supported by the National Natural Science Foundation of China (Grant No. 72192823) and the Zhejiang Key Laboratory of Decision Intelligence (Grant No. 2025E10006). H. Ma (hongma@zju.edu.cn) and W. Zhou (larryzhou@zju.edu.cn) are the corresponding authors.

\bibliographystyle{named}
\bibliography{ijcai26}

\begin{thebibliography}{}

\bibitem[\protect\citeauthoryear{Bello \bgroup \em et al.\egroup }{2016}]{bello2016neural}
Irwan Bello, Hieu Pham, Quoc~V Le, Mohammad Norouzi, and Samy Bengio.
\newblock Neural combinatorial optimization with reinforcement learning.
\newblock {\em arXiv preprint arXiv:1611.09940}, 2016.

\bibitem[\protect\citeauthoryear{Chen \bgroup \em et al.\egroup }{1995}]{exact_alg_chen1995}
Chin-Sheng Chen, Shen-Ming Lee, and QS~Shen.
\newblock An analytical model for the container loading problem.
\newblock {\em European Journal of operational research}, 80(1):68--76, 1995.

\bibitem[\protect\citeauthoryear{Crainic \bgroup \em et al.\egroup }{2009}]{tabu_crainic2009}
Teodor~Gabriel Crainic, Guido Perboli, and Roberto Tadei.
\newblock Ts2pack: A two-level tabu search for the three-dimensional bin packing problem.
\newblock {\em European Journal of Operational Research}, 195(3):744--760, 2009.

\bibitem[\protect\citeauthoryear{Cui \bgroup \em et al.\egroup }{2025}]{entropy_cov}
Ganqu Cui, Yuchen Zhang, Jiacheng Chen, Lifan Yuan, Zhi Wang, Yuxin Zuo, Haozhan Li, Yuchen Fan, Huayu Chen, Weize Chen, et~al.
\newblock The entropy mechanism of reinforcement learning for reasoning language models.
\newblock {\em arXiv preprint arXiv:2505.22617}, 2025.

\bibitem[\protect\citeauthoryear{Dahmani \bgroup \em et al.\egroup }{2025}]{DAHMANI2025100616}
Nadia Dahmani, Amril Nazir, Ikbal Taleb, and Syed M.~Salman Bukhari.
\newblock Reinforcement learning based intelligent optimisation for bin packing problems: A review.
\newblock {\em Array}, 28:100616, 2025.

\bibitem[\protect\citeauthoryear{Dolgui \bgroup \em et al.\egroup }{2019}]{dolgui2019scheduling}
Alexandre Dolgui, Dmitry Ivanov, Suresh~P Sethi, and Boris Sokolov.
\newblock Scheduling in production, supply chain and industry 4.0 systems by optimal control: fundamentals, state-of-the-art and applications.
\newblock {\em International journal of production research}, 57(2):411--432, 2019.

\bibitem[\protect\citeauthoryear{Duan \bgroup \em et al.\egroup }{2019}]{table_2018}
Lu~Duan, Haoyuan Hu, Yu~Qian, Yu~Gong, Xiaodong Zhang, Yinghui Xu, and Jiangwen Wei.
\newblock A multi-task selected learning approach for solving 3d flexible bin packing problem.
\newblock In {\em International Conference on Autonomous Agents and Multiagent Systems}, 2019.

\bibitem[\protect\citeauthoryear{Feng and Yang}{2025}]{feng2025sorrel}
Shengyu Feng and Yiming Yang.
\newblock Sorrel: Suboptimal-demonstration-guided reinforcement learning for learning to branch.
\newblock In {\em Proceedings of the AAAI Conference on Artificial Intelligence}, volume~39, pages 11212--11220, 2025.

\bibitem[\protect\citeauthoryear{Fowler \bgroup \em et al.\egroup }{1981}]{np_hard_fowler1981optimal}
Robert~J Fowler, Michael~S Paterson, and Steven~L Tanimoto.
\newblock Optimal packing and covering in the plane are np-complete.
\newblock {\em Information processing letters}, 12(3):133--137, 1981.

\bibitem[\protect\citeauthoryear{Fu \bgroup \em et al.\egroup }{2021}]{fu2021generalize}
Zhang-Hua Fu, Kai-Bin Qiu, and Hongyuan Zha.
\newblock Generalize a small pre-trained model to arbitrarily large tsp instances.
\newblock In {\em Proceedings of the AAAI conference on artificial intelligence}, volume~35, pages 7474--7482, 2021.

\bibitem[\protect\citeauthoryear{Goyal and Deng}{2020}]{table_2020_1}
Ankit Goyal and Jia Deng.
\newblock Packit: A virtual environment for geometric planning.
\newblock In {\em International Conference on Machine Learning}, pages 3700--3710. PMLR, 2020.

\bibitem[\protect\citeauthoryear{Hu \bgroup \em et al.\egroup }{2017}]{table_2017}
Haoyuan Hu, Xiaodong Zhang, Xiaowei Yan, Longfei Wang, and Yinghui Xu.
\newblock Solving a new 3d bin packing problem with deep reinforcement learning method.
\newblock {\em arXiv preprint arXiv:1708.05930}, 2017.

\bibitem[\protect\citeauthoryear{Hu \bgroup \em et al.\egroup }{2020}]{table_2020_2}
Ruizhen Hu, Juzhan Xu, Bin Chen, Minglun Gong, Hao Zhang, and Hui Huang.
\newblock Tap-net: transport-and-pack using reinforcement learning.
\newblock {\em ACM Transactions on Graphics (TOG)}, 39(6):1--15, 2020.

\bibitem[\protect\citeauthoryear{Huang \bgroup \em et al.\egroup }{2025}]{huang2025balancing}
Shihong Huang, Hongyi Zhu, Hongyuan Wang, Kanglin Liu, Xiaohang Liu, and Zhi-Hai Zhang.
\newblock Balancing the trade-off between efficiency and equity in a stochastic emergency supplies allocation problem.
\newblock {\em Applied Mathematical Modelling}, page 116242, 2025.

\bibitem[\protect\citeauthoryear{Jiang \bgroup \em et al.\egroup }{2021a}]{table_2021_1}
Yuan Jiang, Zhiguang Cao, and Jie Zhang.
\newblock Learning to solve 3-d bin packing problem via deep reinforcement learning and constraint programming.
\newblock {\em IEEE transactions on cybernetics}, 53(5):2864--2875, 2021.

\bibitem[\protect\citeauthoryear{Jiang \bgroup \em et al.\egroup }{2021b}]{table_2021solving}
Yuan Jiang, Zhiguang Cao, and Jie Zhang.
\newblock Solving 3d bin packing problem via multimodal deep reinforcement learning.
\newblock In {\em International Conference on Autonomous Agents and Multiagent Systems}, 2021.

\bibitem[\protect\citeauthoryear{Khalil \bgroup \em et al.\egroup }{2017}]{khalil2017learning}
Elias Khalil, Hanjun Dai, Yuyu Zhang, Bistra Dilkina, and Le~Song.
\newblock Learning combinatorial optimization algorithms over graphs.
\newblock {\em Advances in neural information processing systems}, 30, 2017.

\bibitem[\protect\citeauthoryear{Laterre \bgroup \em et al.\egroup }{2018}]{table_2019}
Alexandre Laterre, Yunguan Fu, Mohamed~Khalil Jabri, Alain-Sam Cohen, David Kas, Karl Hajjar, Torbjorn~S Dahl, Amine Kerkeni, and Karim Beguir.
\newblock Ranked reward: Enabling self-play reinforcement learning for combinatorial optimization.
\newblock In {\em Neural Information Processing Systems}, 2018.

\bibitem[\protect\citeauthoryear{Li \bgroup \em et al.\egroup }{2020}]{table_2020_3}
Dongda Li, Changwei Ren, Zhaoquan Gu, Yuexuan Wang, and Francis Lau.
\newblock Solving packing problems by conditional query learning.
\newblock 2020.

\bibitem[\protect\citeauthoryear{Li \bgroup \em et al.\egroup }{2022}]{li2022one}
Dongda Li, Zhaoquan Gu, Yuexuan Wang, Changwei Ren, and Francis~CM Lau.
\newblock One model packs thousands of items with recurrent conditional query learning.
\newblock {\em Knowledge-Based Systems}, 235:107683, 2022.

\bibitem[\protect\citeauthoryear{Li \bgroup \em et al.\egroup }{2024}]{li2024mathematical}
Yantong Li, Shengjie Wang, Shanshan Zhou, and Zheng Wang.
\newblock A mathematical formulation and a tabu search heuristic for the joint vessel-uav routing problem.
\newblock {\em Computers \& Operations Research}, 169:106723, 2024.

\bibitem[\protect\citeauthoryear{Li \bgroup \em et al.\egroup }{2025}]{li2025collaborative}
Yantong Li, Shengjie Wang, Hairui Sun, and Shanshan Zhou.
\newblock Collaborative vessel--unmanned aerial vehicle routing for time-window-constrained offshore parcel delivery.
\newblock {\em Transportation Research Part C: Emerging Technologies}, 178:105189, 2025.

\bibitem[\protect\citeauthoryear{Liu \bgroup \em et al.\egroup }{2025}]{liu2025cpgd}
Zongkai Liu, Fanqing Meng, Lingxiao Du, Zhixiang Zhou, Chao Yu, Wenqi Shao, and Qiaosheng Zhang.
\newblock Cpgd: Toward stable rule-based reinforcement learning for language models.
\newblock {\em arXiv preprint arXiv:2505.12504}, 2025.

\bibitem[\protect\citeauthoryear{Loshchilov and Hutter}{2019}]{loshchilov2017decoupled}
Ilya Loshchilov and Frank Hutter.
\newblock Decoupled weight decay regularization.
\newblock In {\em International Conference on Learning Representations}, 2019.

\bibitem[\protect\citeauthoryear{Luo \bgroup \em et al.\egroup }{2023}]{luo2023neural}
Fu~Luo, Xi~Lin, Fei Liu, Qingfu Zhang, and Zhenkun Wang.
\newblock Neural combinatorial optimization with heavy decoder: Toward large scale generalization.
\newblock {\em Advances in Neural Information Processing Systems}, 36:8845--8864, 2023.

\bibitem[\protect\citeauthoryear{Martello \bgroup \em et al.\egroup }{2000}]{exact_alg_martello2000three}
Silvano Martello, David Pisinger, and Daniele Vigo.
\newblock The three-dimensional bin packing problem.
\newblock {\em Operations research}, 48(2):256--267, 2000.

\bibitem[\protect\citeauthoryear{Ng \bgroup \em et al.\egroup }{1999}]{ng1999policy}
Andrew~Y Ng, Daishi Harada, and Stuart Russell.
\newblock Policy invariance under reward transformations: Theory and application to reward shaping.
\newblock In {\em Icml}, volume~99, pages 278--287. Citeseer, 1999.

\bibitem[\protect\citeauthoryear{Schrijver and others}{2003}]{schrijver2003combinatorial}
Alexander Schrijver et~al.
\newblock {\em Combinatorial optimization: polyhedra and efficiency}, volume~24.
\newblock Springer, 2003.

\bibitem[\protect\citeauthoryear{Sutton}{1988}]{drl_1988}
Richard~S Sutton.
\newblock Learning to predict by the methods of temporal differences.
\newblock {\em Machine learning}, 3(1):9--44, 1988.

\bibitem[\protect\citeauthoryear{Veres and Moussa}{2019}]{veres2019deep}
Matthew Veres and Medhat Moussa.
\newblock Deep learning for intelligent transportation systems: A survey of emerging trends.
\newblock {\em IEEE Transactions on Intelligent transportation systems}, 21(8):3152--3168, 2019.

\bibitem[\protect\citeauthoryear{Wang \bgroup \em et al.\egroup }{2025a}]{table_2025bin}
Baoying Wang, Zhaohui Lin, Weijie Kong, and Huixu Dong.
\newblock Bin packing optimization via deep reinforcement learning.
\newblock {\em IEEE Robotics and Automation Letters}, 2025.

\bibitem[\protect\citeauthoryear{Wang \bgroup \em et al.\egroup }{2025b}]{entropy_8020}
Shenzhi Wang, Le~Yu, Chang Gao, Chujie Zheng, Shixuan Liu, Rui Lu, Kai Dang, Xionghui Chen, Jianxin Yang, Zhenru Zhang, et~al.
\newblock Beyond the 80/20 rule: High-entropy minority tokens drive effective reinforcement learning for llm reasoning.
\newblock In {\em Advances in Neural Information Processing Systems}, 2025.

\bibitem[\protect\citeauthoryear{Wu \bgroup \em et al.\egroup }{2010}]{ga_wu2010three}
Yong Wu, Wenkai Li, Mark Goh, and Robert De~Souza.
\newblock Three-dimensional bin packing problem with variable bin height.
\newblock {\em European journal of operational research}, 202(2):347--355, 2010.

\bibitem[\protect\citeauthoryear{Wu \bgroup \em et al.\egroup }{2023}]{wu2023machine_Literature_review}
Wenjie Wu, Changjun Fan, Jincai Huang, Zhong Liu, and Junchi Yan.
\newblock Machine learning for the multi-dimensional bin packing problem: Literature review and empirical evaluation.
\newblock {\em arXiv preprint arXiv:2312.08103}, 2023.

\bibitem[\protect\citeauthoryear{Wu \bgroup \em et al.\egroup }{2024}]{wu2024neural}
Xuan Wu, Di~Wang, Lijie Wen, Yubin Xiao, Chunguo Wu, Yuesong Wu, Chaoyu Yu, Douglas~L Maskell, and You Zhou.
\newblock Neural combinatorial optimization algorithms for solving vehicle routing problems: A comprehensive survey with perspectives.
\newblock {\em arXiv preprint arXiv:2406.00415}, 2024.

\bibitem[\protect\citeauthoryear{Xiong \bgroup \em et al.\egroup }{2024}]{xiong2024gopt}
Heng Xiong, Changrong Guo, Jian Peng, Kai Ding, Wenjie Chen, Xuchong Qiu, Long Bai, and Jianfeng Xu.
\newblock Gopt: Generalizable online 3d bin packing via transformer-based deep reinforcement learning.
\newblock {\em IEEE Robotics and Automation Letters}, 9(11):10335--10342, 2024.

\bibitem[\protect\citeauthoryear{Yang \bgroup \em et al.\egroup }{2023}]{yang2023heuristics}
Shuo Yang, Shuai Song, Shilei Chu, Ran Song, Jiyu Cheng, Yibin Li, and Wei Zhang.
\newblock Heuristics integrated deep reinforcement learning for online 3d bin packing.
\newblock {\em IEEE Transactions on Automation Science and Engineering}, 21(1):939--950, 2023.

\bibitem[\protect\citeauthoryear{Zhang and Shuai}{2024}]{zhang2024online}
Jiawei Zhang and Tianping Shuai.
\newblock Online three-dimensional bin packing: A drl algorithm with the buffer zone.
\newblock {\em Foundations of Computing and Decision Sciences}, 49(1):63--74, 2024.

\bibitem[\protect\citeauthoryear{Zhang \bgroup \em et al.\egroup }{2021}]{table_2021attend2pack}
Jingwei Zhang, Bin Zi, and Xiaoyu Ge.
\newblock Attend2pack: Bin packing through deep reinforcement learning with attention.
\newblock {\em arXiv preprint arXiv:2107.04333}, 2021.

\bibitem[\protect\citeauthoryear{Zhao \bgroup \em et al.\egroup }{2021a}]{zhao2021online}
Hang Zhao, Qijin She, Chenyang Zhu, Yin Yang, and Kai Xu.
\newblock Online 3d bin packing with constrained deep reinforcement learning.
\newblock In {\em Proceedings of the AAAI Conference on Artificial Intelligence}, volume~35, pages 741--749, 2021.

\bibitem[\protect\citeauthoryear{Zhao \bgroup \em et al.\egroup }{2021b}]{zhao2021learning}
Hang Zhao, Yang Yu, and Kai Xu.
\newblock Learning efficient online 3d bin packing on packing configuration trees.
\newblock In {\em International conference on learning representations}, 2021.

\bibitem[\protect\citeauthoryear{Zhao \bgroup \em et al.\egroup }{2022}]{zhao1}
Hang Zhao, Chenyang Zhu, Xin Xu, Hui Huang, and Kai Xu.
\newblock Learning practically feasible policies for online 3d bin packing.
\newblock {\em Science China Information Sciences}, 65(1):112105, 2022.

\bibitem[\protect\citeauthoryear{Zhao \bgroup \em et al.\egroup }{2023}]{zhao2}
Hang Zhao, Zherong Pan, Yang Yu, and Kai Xu.
\newblock Learning physically realizable skills for online packing of general 3d shapes.
\newblock {\em ACM Transactions on Graphics}, 42(5):1--21, 2023.

\bibitem[\protect\citeauthoryear{Zhao \bgroup \em et al.\egroup }{2024}]{table_2024dynamic}
Anhao Zhao, Tianrui Li, and Liangcai Lin.
\newblock A dynamic multi-modal deep reinforcement learning framework for 3d bin packing problem.
\newblock {\em Knowledge-Based Systems}, 299:111990, 2024.

\bibitem[\protect\citeauthoryear{Zhao \bgroup \em et al.\egroup }{2025}]{zhao4}
Hang Zhao, Juzhan Xu, Kexiong Yu, Ruizhen Hu, Chenyang Zhu, Bo~Du, and Kai Xu.
\newblock Deliberate planning of 3d bin packing on packing configuration trees.
\newblock {\em The International Journal of Robotics Research}, page 02783649251380619, 2025.

\bibitem[\protect\citeauthoryear{Zhou \bgroup \em et al.\egroup }{2024}]{zhou2024efficient}
Peiwen Zhou, Ziyan Gao, Chenghao Li, and Nak~Young Chong.
\newblock An efficient deep reinforcement learning model for online 3d bin packing combining object rearrangement and stable placement.
\newblock In {\em 2024 24th International Conference on Control, Automation and Systems (ICCAS)}, pages 964--969. IEEE, 2024.

\end{thebibliography}

\end{document}